\def\year{2022}\relax
\documentclass[letterpaper]{article} % DO NOT CHANGE THIS
\usepackage{aaai22}  % DO NOT CHANGE THIS
\usepackage{times}  % DO NOT CHANGE THIS
\usepackage{helvet}  % DO NOT CHANGE THIS
\usepackage{courier}  % DO NOT CHANGE THIS
\usepackage[hyphens]{url}  % DO NOT CHANGE THIS
\usepackage{graphicx} % DO NOT CHANGE THIS
\usepackage{natbib}  % DO NOT CHANGE THIS AND DO NOT ADD ANY OPTIONS TO IT
\usepackage{caption} % DO NOT CHANGE THIS AND DO NOT ADD ANY OPTIONS TO IT
\DeclareCaptionStyle{ruled}{labelfont=normalfont,labelsep=colon,strut=off} % DO NOT CHANGE THIS
\usepackage{algorithm}
\usepackage{algorithmic}
\usepackage{amsmath}
\usepackage[switch]{lineno} 
\usepackage{hyperref}
\usepackage{newfloat}
\usepackage{listings}
\floatstyle{ruled}
\newfloat{listing}{tb}{lst}{}
\floatname{listing}{Listing}
\title{ReforesTree: A Dataset for Estimating Tropical Forest Carbon Stock with Deep Learning and Aerial Imagery}

\author {
    Gyri Reiersen\textsuperscript{\rm 1,2},
    David Dao\textsuperscript{\rm 2},
    Björn Lütjens\textsuperscript{\rm 3},
    Konstantin Klemmer\textsuperscript{\rm 4,5},
    Kenza Amara\textsuperscript{\rm 2},
    Attila Steinegger\textsuperscript{\rm 6},
    Ce Zhang\textsuperscript{\rm 2},
    Xiaoxiang Zhu\textsuperscript{\rm 1}
}
\affiliations {
    \textsuperscript{\rm 1} Technical University of Munich \\
    \textsuperscript{\rm 2} ETH Zurich \\
    \textsuperscript{\rm 3} Massachusetts Institute of Technology \\
    \textsuperscript{\rm 4} University of Warwick \\
    \textsuperscript{\rm 5} New York University \\
    \textsuperscript{\rm 6} WWF Switzerland \\
    gyri.reiersen@tum.de, david.dao@inf.ethz.ch
}

\begin{document}
%\linenumbers

\maketitle

\begin{abstract}
Forest biomass is a key influence for future climate, and the world urgently needs highly scalable financing schemes, such as carbon offsetting certifications, to protect and restore forests. Current manual forest carbon stock inventory methods of measuring single trees by hand are time, labour, and cost intensive and have been shown to be subjective. They can lead to substantial overestimation of the carbon stock and ultimately distrust in forest financing. The potential for impact and scale of leveraging advancements in machine learning and remote sensing technologies is promising, but needs to be of high quality in order to replace the current forest stock protocols for certifications. 

In this paper, we present \textbf{ReforesTree}, a benchmark dataset of forest carbon stock in six agro-forestry carbon offsetting sites in Ecuador. Furthermore, we show that a deep learning-based end-to-end model using individual tree detection from low cost RGB-only drone imagery is accurately estimating forest carbon stock within official carbon offsetting certification standards. Additionally, our baseline CNN model outperforms state-of-the-art satellite-based forest biomass and carbon stock estimates for this type of small-scale, tropical agro-forestry sites. We present this dataset to encourage machine learning research in this area to increase accountability and transparency of monitoring, verification and reporting (MVR) in carbon offsetting projects, as well as scaling global reforestation financing through accurate remote sensing.

\end{abstract}

\section{Introduction}

The degradation of the natural world is unprecedented in human history and a key driver of the climate crisis and the Holocene extinction \cite{Ceballos_2018}. Forests play a significant role in the planet’s carbon cycle, directly impacting local and global climate through its biogeophysical effects and as carbon sinks, sequestering and storing carbon through photosynthesis \cite{Griscom11645}.

However, since the year 2000, we have lost 361 million ha of forest cover, equivalent to the size of Europe, mainly in tropical areas \cite{Hansen_2013}. This accounts for 18\% of global anthropogenic emissions and contributes to driving up atmospherical carbon levels \cite{IPCC_land_use_2019}. Forests, especially tropical forests, also provide habitats for 80\% of land-based biodiversity and with the increasing risk and frequency of wildfires, droughts, and extreme weather, forest ecosystems are under severe pressure \cite{Shi_2021}. 

To avoid planetary tipping points \cite{Rockstrom_2009} and maintain a stable and livable climate, mankind urgently need to reduce carbon emissions until 2050 and restore essential ecosystems \cite{IPCC_2021}. Forests and natural carbon sequestration are important climate change mitigation strategies \cite{Canadell_2008} with a biophysical mitigation potential of 5,380 MtCO2 per year on average until 2050 \cite{IPCC_land_use_2019}. 

% FIXME: The paragraph below has no citations at all
Forestry is a large industry and the causes of deforestation are mostly economically driven \cite{FAO2020} \cite{Geist2001WhatDT}. For the last 20 years, major conservation efforts have been underway to mitigate and safeguard against these losses. One of the global financing strategies is carbon offsets \cite{voluntaryMarket}. Initially, it started as the Clean Development Mechanism (CDM) under the Kyoto Protocol, allowing governments and business organizations from industrialized countries to invest in forestry in developing countries by buying carbon credits to offset industrialized emissions \cite{FAO2020} Several other independent bodies have later developed official standards for verifying and certifying carbon offsetting projects, such as the Gold Standard (GS) and the Verified Carbon Standard (VERRA). The certification process for forest carbon offsetting projects is capital and labour intensive, especially due to the high cost of manual monitoring, verification and reporting (MVR) of the forest carbon stock. 

The carbon offsetting market is rapidly increasing and expected to grow by a factor of 100 until 2050 due to high demand and available capital \cite{voluntaryMarket}. However, the main obstacle is limited supply of offsetting projects as forest owners lack upfront capital and market access \cite{Kreibich_2021}.

Recent research investigations \cite{Badgley21, West24188} have shown that the current manual forest carbon stock practices systematically overestimate forestry carbon offsetting projects with up to 29\% of the offsets analyzed, totaling up to 30 million tCO2e (CO\textsubscript{2} equivalents) and worth approximately \$410 million. The overestimation was identified to come from subjective estimations and modeling of the carbon stock baseline and of the project's additionally and leakage reporting. There is thus a need for higher quality carbon offsetting protocols and higher transparency and accountability of the MVR of these projects \cite{haya2020}. 

There are three key aspects that are important for the use of remote sensing in MVR of forest carbon stock. One aspect is financial; using available and accessible technology and sensors to lower the cost and upfront capital requirements for forest owners to get certified, especially in low and middle-income countries. The second aspect is reducing subjectivity in estimating carbon stock and increasing trustworthiness and transparency in the carbon offsetting certification protocols. And lastly, the solutions need to be scalable due to the urgency of financing forest restoration, especially in tropical regions.

Various verification bodies, new ventures, and academia are currently developing remote sensing technologies to automate parts of the certification process of forestry carbon offsetting projects \cite{rs12111824, daogainforest}. Satellite imagery is increasing in quality and availability and, combined with state-of-the-art deep learning and lidar, promises to soon map every tree on earth \cite{Hanan2020} and to enable forest aboveground biomass and carbon to be estimated at scale \cite{Saatchi9899, Santoro2021}. Compared to current manual estimates, these advancements reduce time and cost and increase transparency and accountability, thus lowering the threshold for forest owners and buyers to enter the market \cite{lut2019}. Nevertheless, these algorithms risk additionally contributing to the systematic overestimation of carbon stocks, not reducing it, and are not applicable for small-scale forests, below 10,000 ha \cite{White_2018}, \cite{GWF}.

Accurately estimating forest carbon stock, especially for small scale carbon offset projects, presents several interesting machine learning challenges, such as high variance of species and occlusion of individual tree crowns. There are many promising approaches, such as hyperspectral species classification \cite{schiefer20}, lidar-based height measurements \cite{Ganz_2019} and individual tree crown segmentation across sites \cite{WEINSTEIN2020101061}. However, these applications have been developed mainly on datasets from temperate forests and, to the knowledge of the authors, there is no publicly available dataset of tropical forests with both aerial imagery and ground truth field measurements.

\begin{figure}
\begin{center}
\centerline{\includegraphics[width=1\columnwidth]{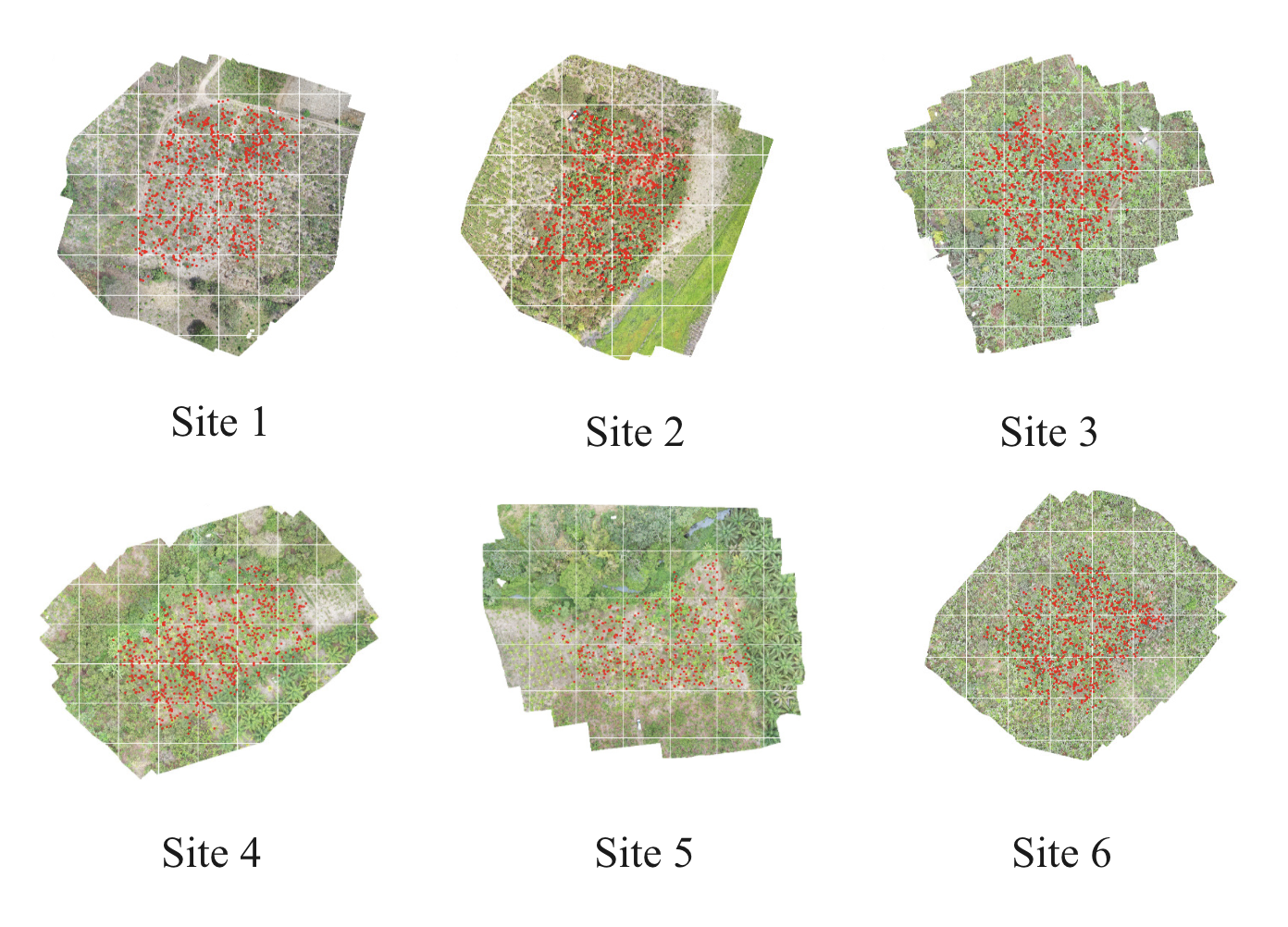}}
\caption{Drone imagery of each site of the ReforesTree dataset with a resolution of 2cm/px. The red dots are the locations of the trees measured in field surveys, plotted to make clear that the coverage of drone images were larger than the field measured area.}
\label{all_sites}
\end{center}
\end{figure}

Here, we present \textbf{ReforesTree}, a dataset of six tropical agroforestry reforestation project sites with individual tree crown bounding boxes of over 4,600 trees matched with their respective diameter at breast height (DBH), species, species group, aboveground biomass (AGB), and carbon stock. This dataset represents ground truth field data mapped with low-cost, high-resolution RGB drone imagery to be used to train new models for carbon offsetting protocols and for benchmark existing models.

To summarize, with \textbf{ReforestTree}, we contribute the following: 1)~the first publicly available dataset of tropical agro-forestry containing both ground truth field data matched with high resolution RGB drone imagery at the individual tree level and 2)~a methodology for reducing the current overestimation of forest carbon stock through deep learning and aerial imagery for carbon offsetting projects.

\section{Related Work}
\subsection{Deep Learning for Remote Sensing}
In recent years, deep learning (DL), and especially deep convolutional neural networks (CNN) are increasing in popularity for image analysis in the remote-sensing community \cite{MA2019166}, \cite{Zhu_2017}. With the increase in computation power, larger datasets, transfer learning, and breakthroughs in network architecture, DL models have outperformed conventional image processing methods in several image tasks such as land use and land cover (LULC) classification, segmentation and detection. Examples of deep supervised learning in remote sensing are the prediction of wildfires \cite{yang2021predicting}, detection of invasive species \cite{Bjorck_2021}. CNNs offer feature extraction capabilities in recognizing patterns in both spatial and temporal data, even with low resolution inputs. With recent advances in meta and few shot learning these models can be trained and generalized on larger datasets and fine-tuned for local variance.

\subsection{Manual Forest Inventory}
\begin{figure}
\begin{center}
\centerline{\includegraphics[width=1\columnwidth]{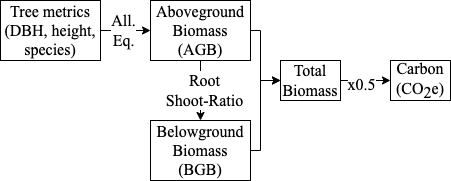}}
\caption{The standard procedure for calculating the correct amount of carbon offsets to be certified for a reforestation project. The tree metrics are collected from manual forest inventory.}
\label{carbon_stock}
\end{center}
\end{figure}

The standardized forest carbon stock inventory consists of manually measuring and registering sample trees of a project site. Tree metrics such as diameter at breast height (DBH), height, and species are then put through scientifically developed regression models called allometric equations to calculate the aboveground biomass (AGB) as seen in Figure \ref{carbon_stock}. The total biomass of a forest is the total AGB added with the below-ground biomass (BGB), calculated using a root-to-shoot ratio specific to the forest type and region \cite{Haozhi_2021}. 

The procedure how to calculate the correct amount of carbon offsets (CO\textsubscript{2}e) to be certified for a project is standardized through  \cite{BioFund} as shown in Figure \ref{carbon_stock}. The (CO\textsubscript{2}e), also known as the baseline forest carbon stock, is equivalent of the total biomass divided by two. Despite being prone to error propagation \cite{Petro_2012, Malhi_2003} and shown to systematically overestimate carbon stock \cite{Badgley21}, this is currently the standardized forest inventory method for certification of forestry projects.

\subsection{Related Methods and Models}
The following are three types of methods to estimate forest carbon stock remotely, adapted from \cite{Sun_2019}; 1)~inventory-based models, based on national and regional forest inventories and regression models, are known to overestimate due to over-representations of dense commercial forests in the data, \cite{GWF}. 2)~Satellite-based models leveraging datasets from optical remote sensing, synthetic aperture radar satellites (SAR), and lidar (LiDAR) to create global aboveground biomass and carbon maps \cite{Santoro2021, Saatchi9899, Spawn2020D}. 3)~Ecosystem-based models using topography, elevation, slope, aspect, and other environmental factors to construct statistical models and quantitatively describe the process of forest carbon cycle to estimate forest carbon stock\cite{Haozhi_2021}.

The most scalable and affordable of these methods are, evidently, satellites-based models. Nevertheless, these models and global maps are yet to estimate carbon stock at local scale and provide accurate estimates of highly heterogeneous and dense forest areas due to their low resolution of 30-300m \cite{Shataee}. An individual tree-based model that takes the individual overstory trees into account can provide this accuracy, especially if fused with geostatistical and satellite data.

In recent years, researchers have achieved high accuracy for standard forestry inventory tasks such as individual tree crown detection \cite{weinstein2019individual}, lidar-based height estimation \cite{Ganz_2019}, and species classification \cite{Miyoshi_2020, schiefer20, MAYRA2021}, using deep learning models and aerial imagery. This shows high potential for combining high-resolution imagery with deep learning models as a method for accurate carbon stock estimation for small-scale reforestation projects \cite{Sun_2019}. 

As most tropical forests are situated in low to middle income countries, without access to hyperspectral, lidar and other more advanced sensors, the models need to be developed using available technologies. A trade-off for accuracy and data availability is basic high-resolution RGB drone imagery. Drone imagery (1-3cm/px resolution), combined with CNN, has previously been used to directly estimate biomass and carbon stock in individual mangrove trees \cite{jones2020} or indirectly by detecting species or tree metrics such as DBH or height \cite{naafalt18, Omasa2003}, achieving an accuracy similar to manual field measurements. And by leveraging multi-fusion approaches \cite{Du2020, Zhang2010}, e.g. combining low-resolution satellite, high-resolution drone imagery, and field measurements and contextual ecological or topological data, and multi-task learning \cite{Crawshaw2020}, e.g. tree metrics and carbon storage factors as auxiliary tasks, these models can replace and scale the existing manual forest inventory.

There are several datasets for tree detection and classification from drone imagery such as the NEON dataset \cite{Weinstein2020}, or the Swedish Forest Agency mainly from temperate forests from the US or Europe. To our knowledge, there are no publicly available datasets including both field measurements and drone imagery of heterogeneous tropical forests. 

\section{Dataset and Method}
The \textbf{ReforesTree} dataset consists of six agro-forestry sites in the central coastal region of Ecuador. The sites are of dry tropical forest type and eligible for carbon offsetting certification with forest inventory done and drone imagery captured in 2020. See Table 1 for information on each site. 

\begin{table}
\label{project-table}
\begin{center}
\begin{small}
\begin{sc}
\begin{tabular}{c c c c c c}
\hline
Site & No. of & No. of & Site & total & total\\
no. & Trees & Species & Area & AGB & CO2e\\

\hline
1 & 743 & 18 & 0.51 & 8  & 5\\ 
2 & 929 & 22 & 0.62 & 15 & 9\\ 
3 & 789 & 20 & 0.48 & 10 & 6\\ 
4 & 484 & 12 & 0.47 & 5  & 3\\ 
5 & 872 & 14 & 0.56 & 15 & 9\\ 
6 & 846 & 16 & 0.53 & 12 & 7\\

\hline
total & 4463 & 28 & 3.17 &  66 & 40\\ 
\hline
\end{tabular}
\end{sc}
\end{small}
\end{center}
\caption{Overview of the six project sites in Ecuador, as gathered in field measurements. Aboveground biomass (AGB) is measured in metric tons and area in hectares.}
\end{table}

\subsection{Forest Inventory Data and Drone Imagery}
Field measurements were done by hand for all live trees and bushes within the site boundaries and include GPS location, species, and diameter at breast height (DBH) per tree. Drone imagery was captured in 2020 by an RGB camera from a Mavic 2 Pro drone with a resolution of 2cm per pixel. Each site is around 0.5 ha, mainly containing banana trees (Musaceae) and cacao plants (Cacao), planted in 2016-2019.

\begin{gather} \label{eq1}
% the log_10 line is the original, the one below was converted by hand
% log_{10}AGB_{fruit} = -0.834 + 2.223 (log_{10}DBH)
AGB_{fruit} = 0.1466 * DBH^{2.223} \\
\label{eq2}
AGB_{musacea} = 0.030 * DBH^{2.13}  \\
\label{eq3}
AGB_{cacao} = 0.1208 * DBH^{1.98}  \\
\label{eq4}
AGB_{timber} = 21.3 - 6.95*DBH + 0.74*DBH^{2}
\end{gather}

The aboveground biomass (AGB) is calculated using published allometric equations for tropical agro-forestry, namely Eq.\ref{eq1} for fruit trees, including citrus fruits \cite{segura2006}, Eq.\ref{eq2} banana trees \cite{van2002}, Eq.\ref{eq3} for cacao  \cite{Yuliasmara2009}, and Eq.\ref{eq4} for shade trees (timber) \cite{Brown1992}. These are commonly used in global certification standards. The carbon stock is calculated through the standard forest inventory methodology using a root-to-shoot ratio of 22\%, which is standard for dry tropical reforestation sites \cite{MA2019166}.

\subsection{Data Processing and Method}

The raw data is processed in several steps as seen in Figure 3. The goal of this process is to have a machine learning ready dataset that consists of matched drone image of an individual tree with the trees labels, such as AGB value. All the drone images have been cropped to fit tightly the boundaries of the field measured areas. The details of this cropping process, and the code repository, are in the Appendix.

\begin{figure}
\begin{center}
\centerline{\includegraphics[width=1\columnwidth]{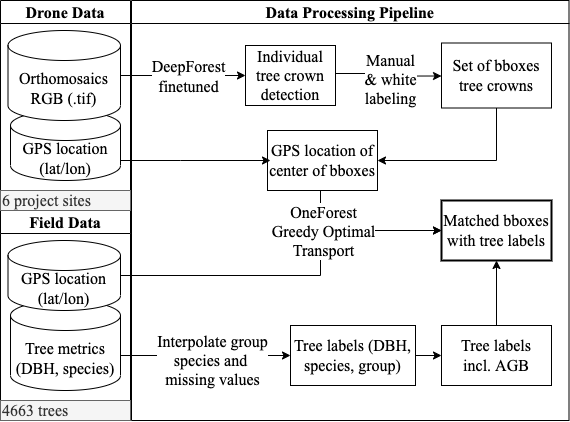}}
\caption{The raw data and data processing pipeline for the ReforesTree dataset, resulting in labels matched to bounding boxes per tree.}
\label{data_pipeline}
\end{center}
\end{figure}

\begin{figure}
\begin{center}
\centerline{\includegraphics[width=1\columnwidth]{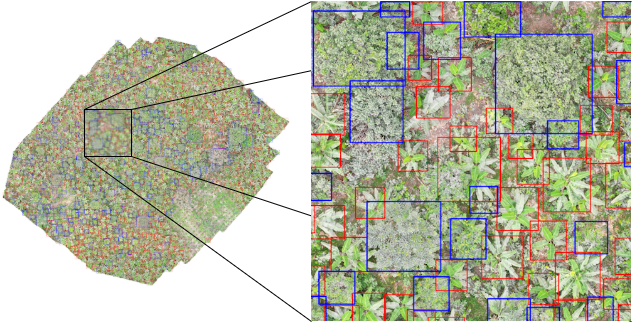}}
\caption{Bounding box annotations per tree, as a result of fine-tuned DeepForest tree crown detection and manual cleaning. Red boxes represent banana trees and blue boxes represent other species.}
\label{bbox}
\end{center}
\end{figure}

Initially the RGB orthomosaics are cut into 4000$\times$4000 tiles and sent through DeepForest, a python package for predicting individual tree crowns from RGB imagery \cite{weinstein2019individual}, fine-tuned on some manually labelled bounding boxes from the sites. Afterwards, the bounding boxes containing more than 80\% white were filtered out, e.g. bounding boxes lying on the border of the drone imagery, and manually labeled to banana and non-banana, due to the easily recognizable characteristics of banana trees, resulting in clear bounding boxes of all trees as shown in Figure~\ref{bbox}. 

To fuse the tree information extracted from the ground measurements with the bounding boxes of the trees detected, we used OneForest, a recent machine learning approach for fusing citizen data with drone imagery. To remove noise introduced in both GPS locations, OneForest uses a greedy optimal transport algorithm. This is a known coupling method to map between two GPS positions (center of bounding box from drone imagery and GPS location of tree from field data). Developed by Villani \cite{villani2003topics}, the methods finds the minimal distance between two distributions via a convex linear program optimizing for a matching that moves the mass from one distribution to the other with minimal cost. The cost is usually defined as the euclidean distance or the Kulback-Leibler divergence between the distributions. The optimum, i.e. the minimal distance between the two distributions, is called the Wasserstein metric. 

\subsection{Baseline CNN Model}
With a dataset of matched bounding boxes and tree labels, we fine-tuned a basic pre-trained CNN, ResNet18 \cite{he2015deep} with a mean-square-error loss to estimate individual tree AGB. The results were satisfying despite the simple baseline model, and proves that the individual tree estimation from drone imagery has potential.

Fourteen images were identified as being larger than the expected crown size of a tree, and they were center cropped at 800$\times$800. To preserve the crown size information, the smaller images were zero-padded up to 800$\times$800, before all images were resized to fit the network architecture. 

The dataset has is unbalanced with regards to species, of which 43\% is cacao and 32\% is banana. Additionally, due to the trees being planted between 2016-2019, many of the trees have similar size (e.g. DBH) and half of the trees have DBH between 7-10cm. The training dataset consisted of equal number of samples of species and DBH, and from the different project sites.

\section{Experiments}
With the emerging new biomass maps and forest stock estimation models, we used the \textbf{ReforesTree} dataset to benchmark these maps and compare with our baseline CNN model for AGB estimation. We compared the maps taken from \cite{GWF}, \cite{Spawn2020D}, and \cite{Santoro2021}. The Global Forest Watch's Above-Ground Woody Biomass dataset is a global map of AGB and carbon density at 30m$\times$30m resolution for the year 2000. It is based on more than 700,000 quality-filtered Geoscience Laser Altimeter System (GLAS) lidar observations using machine learning models based on allometric equations for the different regions and vegetation types. The second dataset from \cite{Spawn2020D} is a 300m$\times$300m harmonized map based on overlayed input maps. The input maps were allocated in proportion to the relative spatial extent of each vegetation type using ancillary maps of tree cover and landcover, and a rule-based decision schema. The last, and most recent 100m$\times$100m dataset from \cite{Santoro2021} is obtained by spaceborne SAR (ALOS PALSAR, Envisat ASAR), optical (Landsat-7), lidar (ICESAT) and auxiliary datasets with multiple estimation procedures with a set of biomass expansion and conversion factors following approaches to extend ground estimates of wood density and stem-to-total biomass expansion factors. 

\begin{table}

\label{experiments-table}
\begin{center}
\begin{small}
\begin{sc}
\begin{tabular}{c c c c c c}
\hline
Site & Field & GFW & Spawn & Santoro & Baseline\\
no. & Data & 2019 & 2020 & 2021 & (Ours)\\

\hline
1 & 8  & 99 & 97  & 36 & 7 \\ 
2 & 15 & 108 & 130 & 42 & 8 \\ 
3 & 10 & 36 & 206  & 15 & 15\\ 
4 & 5  & 5  & 102  & 32 & 9 \\ 
5 & 15 & 73 & 352  & 12 & 11\\ 
6 & 12 & 26 & 91  & 72 & 15\\

\hline
tot. & 66 & 331 & 413 & 89 & 65\\ 
\hline
\end{tabular}
\end{sc}
\end{small}
\end{center}
\caption{The benchmark results from comparing different models for estimating AGB with the forest inventory of the ReforesTree sites. All numbers are given as AGB in Mg. GFW is \cite{GWF}, Spawn is \cite{Spawn2020D}, Santoro is \cite{Santoro2021}. All of these three are satellite-based. Lastly, the baseline CNN is our drone-based model.}
\end{table}

As seen in Table 2, all of the available global AGB maps have a tendency to overestimate the ground truth measurements up to a factor of ten. These are not encouraging results showing that these maps are far from being accurate enough to be used in remote sensing of forest carbon stock at a small scale, as is the case for the \textbf{ReforesTree} dataset.

Our baseline model, on the other hand, has a slight tendency of underestimating the biomass. The model has an evident advantage, to be trained on the dataset, but these initial results show promise for the individual tree estimation approach using drone imagery for forest carbon inventory.

\section{Conclusions and Future Work}
We introduce the \textbf{ReforesTree} dataset in hopes of encouraging the fellow machine learning community to take on the challenge of developing low-cost, scalable, trustworthy and accurate solutions for monitoring, verification and reporting of tropical reforestation inventory. We also present an outlined methodology for creating an annotated machine learning dataset from field data and drone imagery, and train a baseline CNN model for individual tree aboveground biomass estimation. This methodology includes a data processing pipeline leveraging a fine-tuned tree crown detection algorithm and an optimal transport matching algorithm for reduction of GPS noise.

The \textbf{ReforesTree} dataset of field measurements and low-cost, high-resolution RGB drone imagery represents the trade-off for accuracy and data availability of remote sensing of forest carbon stock in tropical regions. It can be used to train new or benchmark existing models for MVR of carbon offsetting reforestation protocols. Remote inventory of small scale tropical reforestation projects comes with several ecological challenges, such high biodiversity, level of canopy closure, and topology. This dataset is a start to develop a generalized model that can be fine-tuned on local scale. Future work will investigate ways to improve the methodology and reduce error in the machine learning ready dataset, and increase the explainability to have a trustworthy and transparent model. Additionally, we see further potential in fusing satellite and other available geoecological data layers as well as leveraging the multiple labels available (e.g. DBH, species) as auxiliary tasks in a multitask learning problem. 

As the world is rapidly approaching planetary doom, we need to collaborate across disciplines to implement and scale the climate mitigation strategies available. Restoration of forests is one of our most important climate mitigation strategies. And by reducing the overestimation of carbon offsets, we can allow every man on earth who owns a tree to participate in climate action. Biodiverse and sustainable forestry can provide hope not only the for the machine learning community, but also beyond.

\section{Acknowledgments}
The authors are thankful for the guidance and advice by our academic collaborator (Prof. Dava Newman, Prof. Lynn H Kaack, Prof. Thomas Crowther and the CrowtherLab), non-governmental institutions (BrainForest, WWF Switzerland, Restor), Isabel Hillman, Simeon Max, Microsoft AI for Earth, and support from the local community in Ecuador. 
Lastly, we extend our sincere gratitude to Autumn Nguyen and Sulagna Saha for their significant contributions to this work. Their thorough review process led to substantial improvements in both the manuscript and the underlying codebase. Their detailed technical analysis and implementations have enhanced the robustness and reliability of our research. A comprehensive report of their contributions can be found in our technical documentation: {\url{https://gainforest.substack.com/p/improving-reforestree-correcting}}.

% Use \bibliography{yourbibfile} instead or the References section will not appear in your paper
\bibliography{aaai22_gyri}

%\clearpage
%\appendix

\section{Technical Appendix}
\subsection{Raw data cleaning}
All 28 species were divided into 6 species family groups: banana, cacao, fruit, timber, citrus and other.

The field data was manually collected as a standard manual forest inventory, potentially leading to human errors, missing values and outliers. 

The dataset needed to reflect the ground truth. Therefore it was important not to remove trees from the dataset unnecessarily. All missing DBH values were given a value based on the average DBH of the same species for the year it was planted. Of the 28 species, only 3 species (in total 25 trees) were missing DBH values: 23 lemon (citrus), one balsa (timber), one bariable (other) trees.  These were given DBH values interpolated from the other trees in the same family group and which were planted the same year.

Additionally, 8 banana trees that had DBH values larger than 50cm, which is unrealistically high. Assuming that there was a manual entry mistake, these values were exchanged with the maximum value of the banana trees for the year planted. 

\begin{figure}[ht]
\begin{center}
\centerline{\includegraphics[width=1\columnwidth]{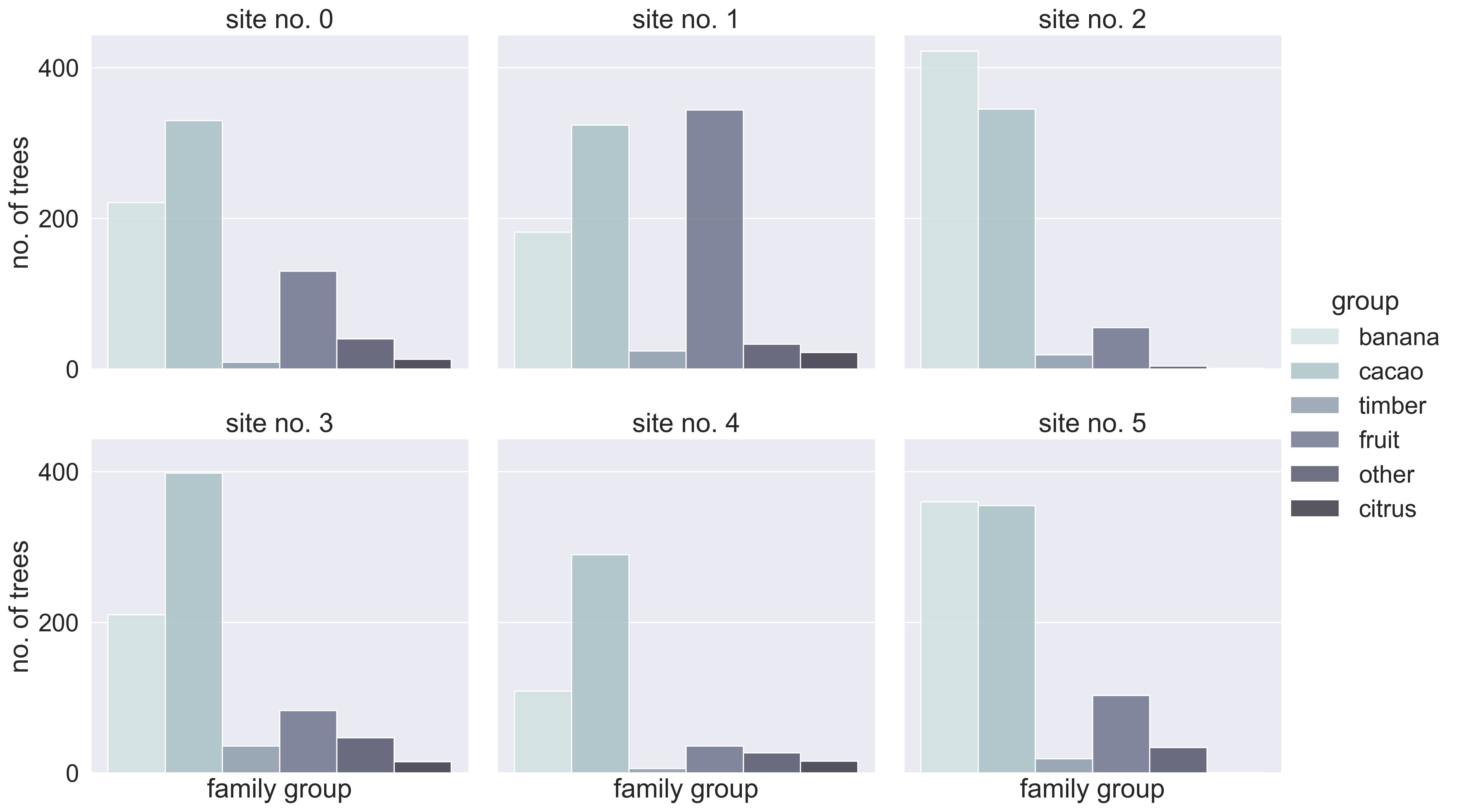}}
\caption{This figure represents the count of species family groups for each of the sites. All sites have trees of all species family groups, but cacao and banana are over represented.}
\label{group_dist}
\end{center}
\end{figure}

\subsection{Aligning Drone Images with Field Boundaries}

A key issue identified in the ReforesTree pipeline was the mismatch between the drone imagery boundaries and the field data boundaries. To address this, we implemented the following steps to align the drone images with the field measurements for the six agroforestry sites. The code for this is in this \href{https://github.com/autumn-yng/reforestree-correction/tree/main}{\textit{\textbf{reforestree-correction}} repository}.

\begin{figure}[ht]
\begin{center}
\centerline{\includegraphics[width=1\columnwidth]{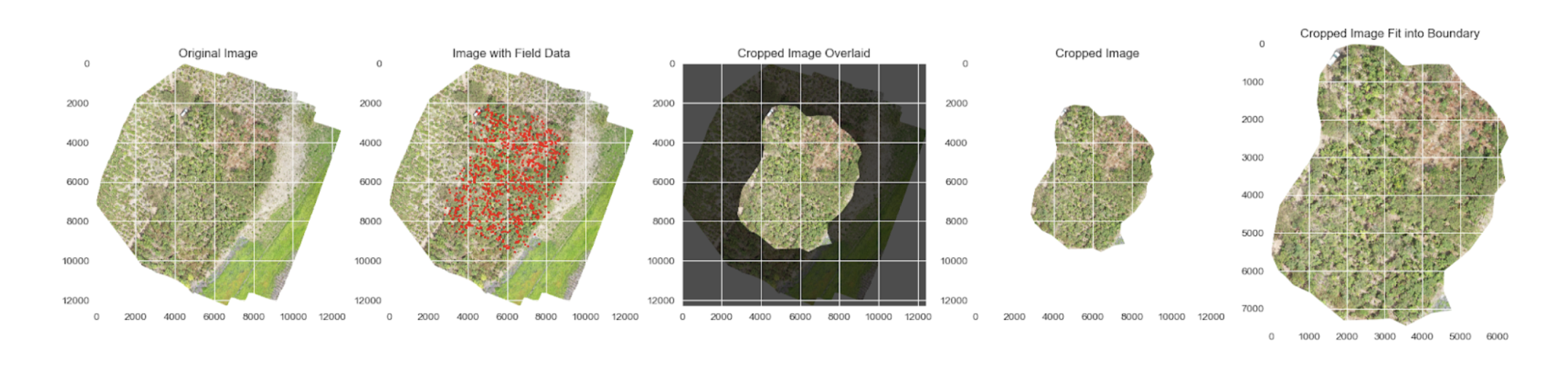}}
\caption{This figure shows the alignment process between the drone images and field boundaries. The field data points (red dots) were used to create the alphashape, which was overlaid onto the drone imagery to crop unnecessary areas and ensure accurate alignment.}
\label{drone_field_alignment}
\end{center}
\end{figure}

\begin{enumerate}
    \item \textbf{GeoDataframe Creation:} We converted the field data, which included the longitude and latitude of each point, into a GeoDataFrame using the geopandas library. This allowed us to create point geometries that were easy to visualize and manipulate. The field data points, visualized as red dots in Figure \ref{drone_field_alignment}, served as the starting reference.
    
    \item \textbf{Boundary Extraction using Alpha Shape:} To capture the boundary of the field data, we used the alphashape library to create a convex hull around the points. By choosing an alpha value of 15000, similar to the value used by \cite{Barenne2022}, we generated a tight boundary around the field data points.
    
    \item \textbf{Overlay and Crop Drone Imagery:} Using the rasterio library, we overlapped the generated alphashape boundary onto the drone imagery (in TIFF format). We then cropped the unnecessary parts of the image, outside the boundary, replacing them with white pixels. This step is illustrated by the transition from the third to fourth images in Figure \ref{drone_field_alignment}.
    
    \item \textbf{Adjusting Image Boundaries:} Finally, after cropping, we identified the bounds of the non-white pixels in the images and adjusted them to ensure they fit a square shape correctly. This was essential for integrating the images into the AGBench library. The final result can be seen in the transition from the fourth to the last image in Figure \ref{drone_field_alignment}.
\end{enumerate}

\subsection{Benchmark of satellite-based AGB maps}
To benchmark the low resolution (LR) satellite-based maps, we fitted it to the high resolution (HR) drone imagery overlapping the GPS coordinates.

\begin{figure}[ht]
\begin{center}
\centerline{\includegraphics[width=1\columnwidth]{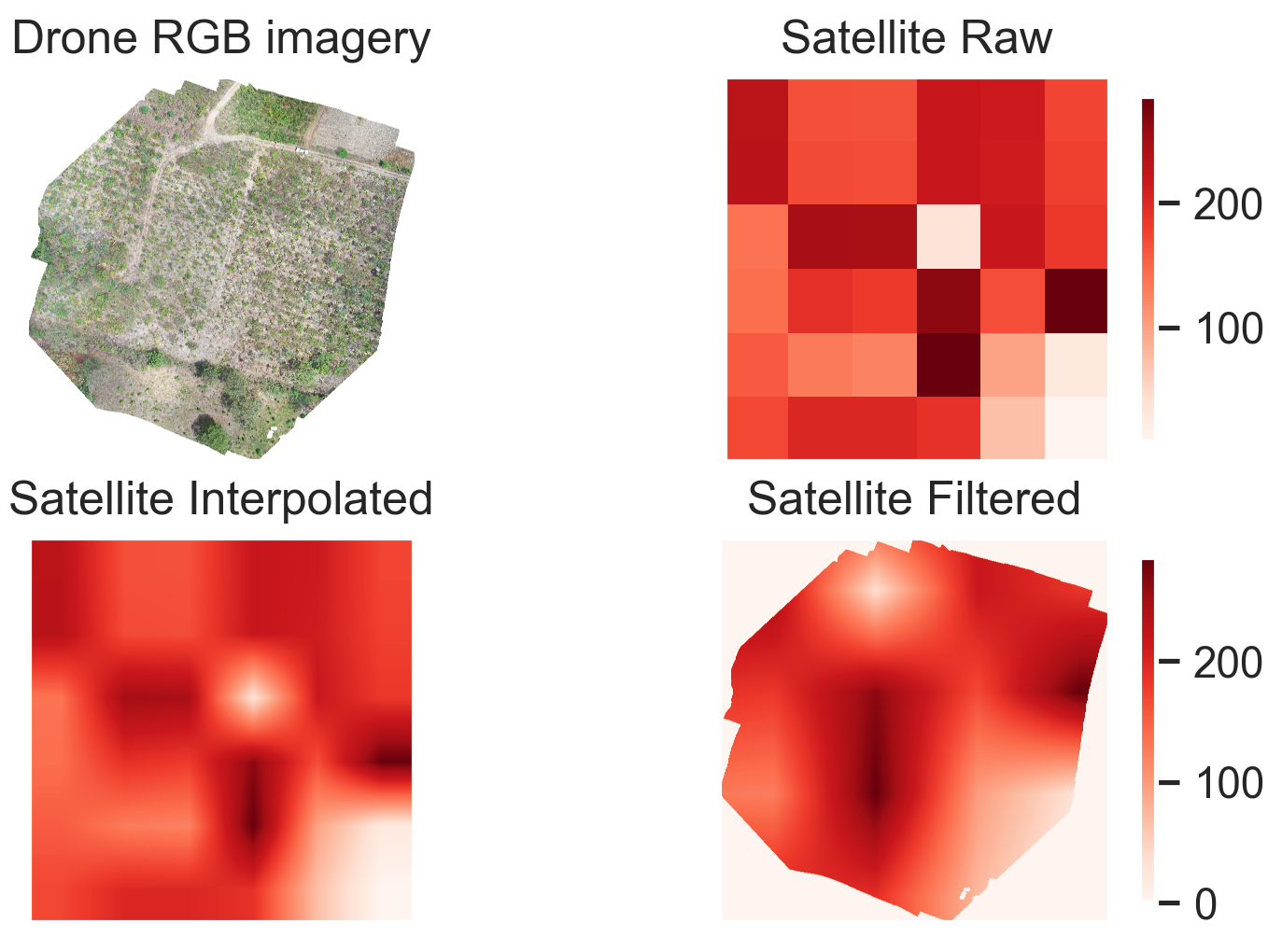}}
\caption{This figure represents the different steps in the benchmark analysis and how we calculated the total AGB amount from the satellite-based maps for the \textbf{ReforesTree} sites. This is taken from site no. 0. The values represented in the image is AGB density.}
\label{satellite}
\end{center}
\end{figure}

The calculation of the total AGB was done in five steps, illustrated in Figure \ref{satellite}
\begin{enumerate}
    \item cropping the LR satellite map with a padding around the polygon of the site to reduce computation intensity (Satellite Raw)
    \item linearly interpolating the values for this map and resize the map with the same HR pixel resolution as the drone imagery (Satellite Interpolated)
    \item cropping the map further fitting with the GPS locations (max/min) of the drone imagery
    \item filtering out the site area by removing all pixels in the satellite-based map, that are outside of the drone imagery, coloured white (Satellite Filtered)
    \item lastly, multiplying the AGB mean density of the filtered map with the project site area to get the total AGB
\end{enumerate}

We analysed the following three maps:

\begin{itemize}
  \item \cite{GWF}: Aboveground Woodly Biomass with 30x30m resolution for the year of 2000.
  \item \cite{Spawn2020D}: Global Aboveground and Belowground Biomass Carbon Density Maps for the Year 2010 with 300x300m resolution.
  \item \cite{santoro2018ggdo}: GlobBiomass - Global Datasets of Forest Biomass with 100x100m resolution for the year 2010.
\end{itemize}

\subsection{Baseline CNN}

We trained the model on a single GPU of the type GeForce RTX 3090. The learning rate used was 1e\textsuperscript{-3}, batch size of 64 for 30 epochs achieving a root square mean loss (RMSE) of 0,1.

\end{document}